\documentclass{article}
\usepackage{spconf,amsmath,amssymb,amsfonts}
\usepackage{graphicx}
\usepackage{booktabs}
\usepackage{multirow}
\usepackage{siunitx}
\usepackage{array}
\usepackage{cite}
\usepackage{placeins}

\newcommand{\fscore}[2]{#1\,\ensuremath{\pm}\,#2}

\title{Tissue Aware Nuclei Detection and Classification Model for Histopathology Images}

\name{Kesi Xu$^{\star}$ \qquad Eleni Chiou$^{\dagger}$ \qquad Ali Varamesh$^{\dagger}$ \qquad Laura Acqualagna$^{\dagger}$ \qquad Nasir Rajpoot$^{\star}$}
\address{
$^{\star}$TIA Centre, Department of Computer Science, University of Warwick, UK \\
$^{\dagger}$GSK, Artificial Intelligence and Machine Learning, London, UK.
}

\begin{document}
\ninept
\maketitle
\begin{abstract}
Accurate nuclei detection and classification are fundamental to computational pathology, yet existing approaches are hindered by reliance on detailed expert annotations and insufficient use of tissue context. We present Tissue-Aware Nuclei Detection (TAND), a novel framework achieving joint nuclei detection and classification using point-level supervision enhanced by tissue mask conditioning. TAND couples a ConvNeXt-based encoder-decoder with a frozen Virchow-2 tissue segmentation branch, where semantic tissue probabilities selectively modulate the classification stream through a novel multi-scale Spatial Feature-wise Linear Modulation (Spatial-FiLM). On the PUMA benchmark, TAND achieves state-of-the-art performance, surpassing both tissue-agnostic baselines and mask-supervised methods. Notably, our approach demonstrates remarkable improvements in tissue-dependent cell types such as epithelium, endothelium, and stroma. To the best of our knowledge, this is the first method to condition per-cell classification on learned tissue masks, offering a practical pathway to reduce annotation burden.
\end{abstract}


\section{Introduction}
Cell detection and classification from Haematoxylin and Eosin stained slides are foundational to computational pathology, enabling quantification of cell-type distributions for biomarker readouts, tumour microenvironment profiling, and patient-level risk models~\cite{graham2024conic}. Many clinical endpoints, such as counts or ratios of immune and epithelial subsets, require reliable cell centres and types rather than full-fidelity instance masks. In practice, to accurately determine a cell type, pathologists first identify the surrounding tissue type as contextual information before making a precise judgement. It is thus crucial to understand the rich prior knowledge of the tissue layer for better nuclei classification.

Instance-level nuclei analysis methods such as HoVerNet and HoVerNeXt~\cite{graham2019hovernet,baumann2024hovernext} typically rely on dense mask supervision and rarely inject tissue context explicitly into per-cell classification. While Co-Seg++~\cite{Co-Seg} explores joint tissue--nuclei segmentation through cross-task prompt fusion in the decoder, it focuses on nuclei instance mask prediction rather than point-based detection and classification. The annotation burden of dense contours has motivated point-based supervision approaches~\cite{qu2019weakly,lin2023point,zhang2023histopoints,yao2024psa}, which leverage rapidly obtainable centre-point labels from tools like QuPath~\cite{bankhead2017qupath}. Meanwhile, pathology foundation models such as Virchow-2~\cite{zimmermann2024virchow2} and self-supervised backbones from the DINO family~\cite{caron2021emerging,oquab2023dinov2} provide robust tissue semantics and dense features. However, DINOv3~\cite{dinov3} remains unexplored for nuclei detection and classification, and tissue-level context has not been explicitly leveraged to condition per-nucleus predictions in point-supervised frameworks.

In this work, we address the task of joint nuclei detection and classification from point-based supervision, introducing an explicit tissue-aware conditioning mechanism within the classification pathway. We present a novel Tissue-Aware Nuclei Detection (TAND) framework that leverages tissue mask conditioning to enhance nuclei classification performance. 

Our key contributions are as follows:  
(1) We explore the role of tissue context in nuclei classification by proposing a multi-scale spatial Feature-wise Linear Modulation (Spatial-FiLM) module that integrates tissue mask prediction features with nuclei detection features to optimise per-nucleus class prediction.  
(2) Motivated by the strong representational capability of DINOv3 in natural image segmentation~\cite{dinov3}, we are the first to adapt DINOv3 for nuclei detection and classification in computational pathology, establishing a robust baseline and methodological foundation for future applications of vision foundation models in digital pathology.  
(3) We propose an alternative to instance-level nuclei segmentation for scenarios where full instance masks are not the primary requirement, achieving competitive nuclei-type generalisation while reducing annotation effort.  

All code for TAND will be made publicly available upon publication.

\section{The Proposed Method}
\subsection{Overview}
\begin{figure*}[t!]
  \centering
  \includegraphics[width=0.97\textwidth]{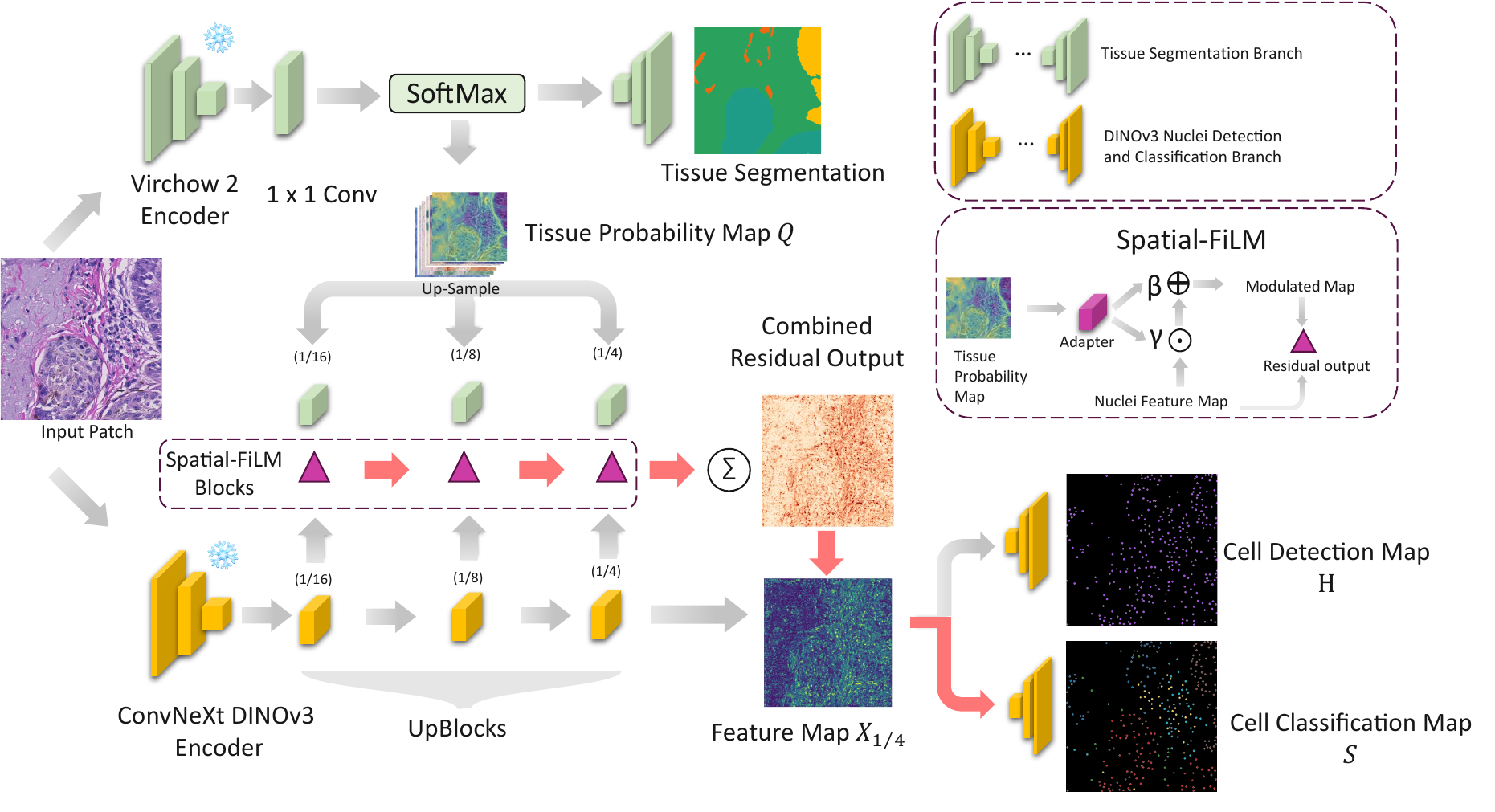}
  \caption{Overall architecture of the proposed Tissue Aware Nuclei Detection and Classification (TAND) framework.
  A frozen Virchow-2 encoder predicts tissue masks probability map. The red arrow shows feature maps modulate only the
  nuclei classification stream via multi-scale Spatial-FiLM at \(1/16\), \(1/8\), and \(1/4\) scales,
  while the nuclei detection path remains unmodulated. }
  \label{fig:arch}
\end{figure*}

We propose Tissue-Aware Nuclei Detection (TAND) model. The proposed model comprises a tissue segmentation branch based on a frozen Virchow-2 encoder with a linear head that outputs tissue probabilities \(\mathbf{Q}\), and a detection--classification branch using a DINOv3-pretrained ConvNeXt encoder with an upsampling decoder that predicts a nuclei-centre detection heatmap \(\mathbf{H}\) and a classification logits map \(\mathbf{S}\in\mathbb{R}^{K\times H\times W}\) over \(K\) nucleus types, where \(H\) and \(W\) denote the spatial height and width dimensions respectively. Tissue probabilities modulate only the classification stream via multi-scale Spatial-FiLM. The overall architecture is shown in Fig.~\ref{fig:arch}.

\subsection{Tissue Segmentation Branch}
We attach a linear head to a frozen Virchow-2 encoder to predict multi-scale tissue logits~\cite{zimmermann2024virchow2}. After a temperature-scaled softmax, tissue probabilities \(\mathbf{Q} \in \mathbb{R}^{T \times H \times W}\), where \(T\) denotes the number of tissue classes, are bilinearly upsampled to match decoder feature scales. The tissue segmentation head is trained using a combination of Soft-Dice loss, cross-entropy (CE) loss and a pixel-wise binary cross-entropy (BCE) loss. Tissue mask prediction is evaluated using per-class Dice scores.

\subsection{DINOv3 Detection-Classification Backbone}
We employ a ConvNeXt-based DINOv3 pretrained encoder and three upsampling blocks to produce \(1/4\)-resolution features \(\mathbf{x}_{1/4}\). Two heads are attached: a single-channel heatmap head \(\mathbf{H}\) for detection and a \(K\)-channel head \(\mathbf{S}\) for classification. The detection path operates on \(\mathbf{x}_{1/4}\) and is left unmodulated by tissue information.

For detection supervision we create Gaussian centre-maps \(\mathbf{Y}\) from annotated point centres at the heatmap resolution (the Gaussian standard deviation \(\sigma\) is tuned on a validation split). The heatmap \(\mathbf{H}\) is supervised with a sigmoid focal loss \(\mathcal{L}_{\mathrm{det}}\) to mitigate pixel class imbalance, since most pixels are not nuclei centres. At inference, centres are recovered by local-maximum suppression on \(\mathbf{H}\).
\subsection{Spatial Feature-wise Linear Modulation}
To effectively incorporate tissue context into cell classification while preserving detection accuracy, we introduce a multi-scale Spatial Feature-wise Linear Modulation (Spatial-FiLM) mechanism that selectively conditions the classification stream on tissue probabilities. Unlike conventional concatenation or attention-based fusion strategies that may dilute fine-grained cellular features, our approach learns tissue-aware affine transformations that adaptively modulate intermediate decoder representations. Specifically, given decoder features \(\mathbf{x}_s \in \mathbb{R}^{C_s \times H_s \times W_s}\) at scale \(s \in \{1/16,\,1/8,\,1/4\}\) (where \(C_s\) denotes the number of channels, \(H_s\) and \(W_s\) the spatial dimensions at scale \(s\)) and upsampled tissue probabilities \(\mathbf{Q}_s \in \mathbb{R}^{T \times H_s \times W_s}\) where \(T\) is the number of tissue types, the Spatial-FiLM module computes spatially-varying modulation parameters through a lightweight convolutional adapter with learnable parameters \(\boldsymbol{\theta}_s\):
\begin{equation}
[\boldsymbol{\gamma}_s,\boldsymbol{\beta}_s] \;=\; \mathcal{F}_s(\mathbf{Q}_s;\boldsymbol{\theta}_s),\qquad
\boldsymbol{\gamma}_s,\boldsymbol{\beta}_s\in\mathbb{R}^{C_s\times H_s\times W_s},
\label{eq:film_params}
\end{equation}
where \(\mathcal{F}_s\) denotes a two-layer convolutional network with hidden dimension \(h_s\). To stabilise training we bound the raw outputs as
\begin{equation}
\tilde{\boldsymbol{\gamma}}_s=\tanh(\boldsymbol{\gamma}_s)\,\eta,\quad \tilde{\boldsymbol{\beta}}_s=\tanh(\boldsymbol{\beta}_s)\,\tfrac{\eta}{2},
\label{eq:bounded_params}
\end{equation}
with \(\eta=0.5\) by default. The FiLM-modulated features are
\begin{equation}
\tilde{\mathbf{x}}_{s}=\mathbf{x}_{s}\odot(1+\tilde{\boldsymbol{\gamma}}_s)+\tilde{\boldsymbol{\beta}}_s,
\label{eq:film_modulation}
\end{equation}
where \(\odot\) denotes element-wise multiplication, and we define the modulation residual \(\Delta_s=\tilde{\mathbf{x}}_{s}-\mathbf{x}_{s}\).
These multi-scale residuals are bilinearly upsampled to \(1/4\) resolution and mapped into the classifier feature space by learned \(1\times1\) projections \(\mathcal{P}_s\) (initialised to zero):
\begin{equation}
\mathbf{x}_{\mathrm{cls}} \;=\; \mathbf{x}_{1/4} \;+\; \sum_{s\in\{1/16,\,1/8,\,1/4\}} \mathcal{P}_s\big(\uparrow_s \Delta_s\big),
\label{eq:multiscale_aggregation}
\end{equation}
where \(\uparrow_s\) denotes bilinear upsampling to the \(1/4\) grid. Zero initialisation of \(\mathcal{P}_s\) ensures the classifier begins as the pretrained baseline and that FiLM only introduces changes as training proceeds. This residual, multi-scale aggregation preserves pretrained representations, enables selective application of tissue-aware modulation, and captures both fine-grained cellular detail and broader tissue context. By restricting FiLM to the classification stream and leaving the detection head unmodified, we explicitly decouple spatial localisation from tissue-aware semantic reasoning.
\begin{figure*}[t!]
  \centering
  \includegraphics[width=0.96\textwidth]{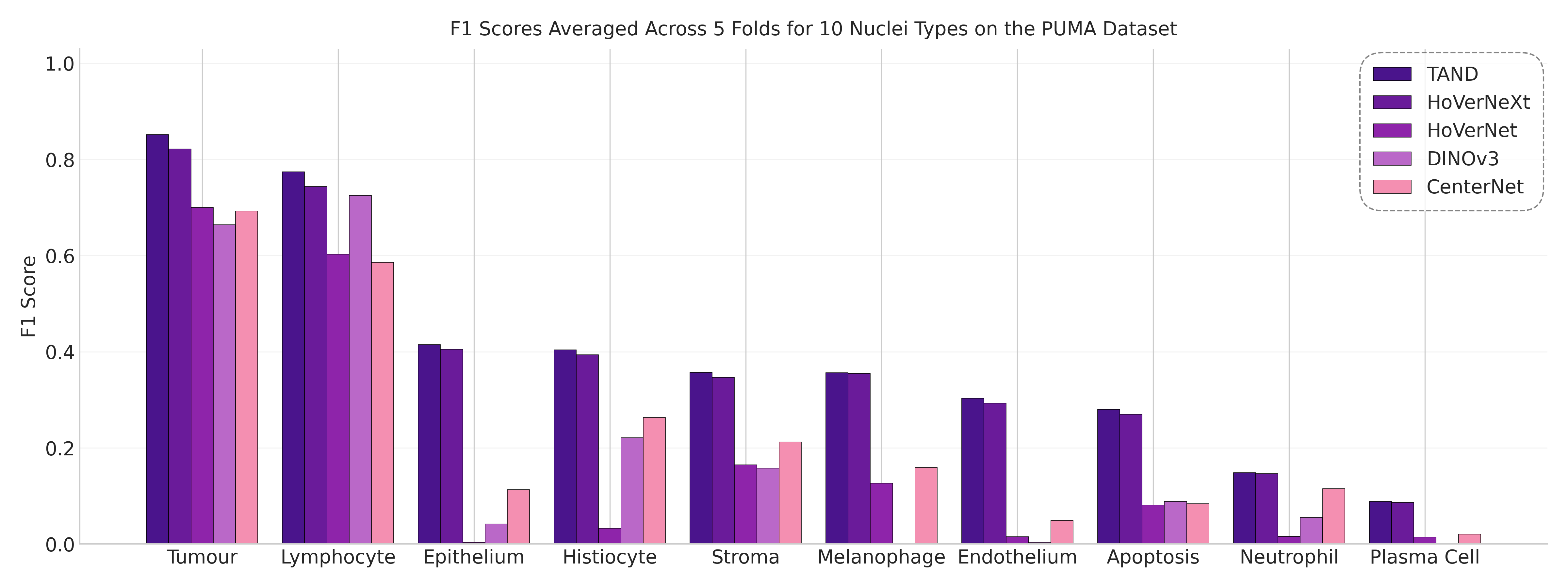}
  \caption{Per-class F1 comparison between the state-of-the-art methods and the proposed TAND model.}
  \label{fig:perclassf1}
\end{figure*}
\subsection{Training Protocol}
We adopt a compact three-stage training curriculum to ensure stability. (i) Only train the tissue segmentation branch, with Virchow 2 encoder frozen, to obtain reliable multi-scale tissue probabilities \(\mathbf{Q}\); then freeze its weights. (ii) With \(\mathbf{Q}\) fixed, train the DINOv3 detection--classification backbone: supervise the heatmap \(\mathbf{H}\) with sigmoid focal loss \(\mathcal{L}_{\mathrm{det}}\), and train classification logits \(\mathbf{S}\) with point-wise cross-entropy \(\mathcal{L}_{\mathrm{cls}}\) sampled at annotated centres plus a local pixel-wise BCE term \(\mathcal{L}_{\mathrm{BCE}}\). (iii) Finally, enable Spatial-FiLM while keeping the tissue segmentation branch frozen; FiLM adapters and the \(1\times1\) projection layers \(\mathcal{P}_s\) are zero-initialised and updated with reduced learning rates so the model departs smoothly from the pretrained baseline. The overall objective is a weighted sum with loss weights \(\lambda_{\mathrm{det}}\), \(\lambda_{\mathrm{cls}}\), and \(\lambda_{\mathrm{BCE}}\):
\begin{equation}
\mathcal{L}=\lambda_{\mathrm{det}}\mathcal{L}_{\mathrm{det}}+\lambda_{\mathrm{cls}}\mathcal{L}_{\mathrm{cls}}+\lambda_{\mathrm{BCE}}\mathcal{L}_{\mathrm{BCE}},
\label{eq:total_loss}
\end{equation}
with weights selected on a validation split. This curriculum yields stable convergence and preserves localisation performance while enabling the classifier to leverage tissue context.

\section{Experiments}
\subsection{Dataset and Protocol} 

We evaluate on the public Panoptic Segmentation of Nuclei and Tissue in Advanced Melanoma (PUMA) dataset~\cite{schuiveling2024puma}. The available training set comprises 206 ROIs (103 primary / 103 metastatic). Following the PUMA benchmark, detection performance is measured by centre matching: for each ground-truth nucleus we search for the highest-scoring prediction within a radius of 15 pixels ($\approx$3.3\,\si{\micro\meter}); matched predictions are removed to enforce one-to-one assignment and remaining predictions are treated as false positives. Detection precision, recall and F1 are computed from these assignments. Classification performance is computed on the matched pairs and reported as per-class F1 and macro-F1. 

As the official PUMA evaluation employs a hidden test set, our results are not directly comparable to the challenge leaderboard. To ensure a fair comparison of model architectures, we deliberately exclude auxiliary optimisation techniques such as test-time augmentation, ensembling, and advanced data augmentation. This controlled setup isolates the intrinsic performance of different architectures. We conduct stratified 5-fold cross-validation with all reported metrics representing mean values across folds. 

All images are $1024 \times 1024$ pixel ROI-level patches acquired at 40$\times$ magnification (0.22\,\si{\micro\meter}/px). The dataset is randomly split into 5 stratified folds preserving the distribution of primary and metastatic samples, where each fold allocates approximately 80\% of ROIs for training and 20\% for testing.

\begin{table}[t]
  \centering
  \caption{Five-fold performance on the PUMA nuclei benchmark.}
  \label{tab:table1}
  \begin{tabular}{l>{\centering\arraybackslash}p{2.5cm}>{\centering\arraybackslash}p{2.5cm}}
    \toprule
    Model & Detection F1 (mean $\pm$ std) & Classification macro-F1 (mean $\pm$ std) \\
    \midrule
    CenterNet~\cite{centernet}  & $\fscore{0.598}{0.079}$ & $\fscore{0.230}{0.032}$ \\
    DINOv3-Unet ~\cite{dinov3}     & $\fscore{0.614}{0.088}$ & $\fscore{0.196}{0.043}$ \\
    HoVerNet~\cite{graham2019hovernet}   & $\fscore{0.616}{0.033}$ & $\fscore{0.175}{0.009}$ \\
    HoVerNeXt~\cite{baumann2024hovernext}  & $\fscore{0.724}{0.026}$ & $\fscore{0.374}{0.022}$ \\
    TAND & $\mathbf{0.734 \pm 0.026}$ & $\mathbf{0.398 \pm 0.024}$ \\
    \bottomrule
  \end{tabular}
\end{table}

\begin{figure*}[t!]
  \centering
  \includegraphics[width=0.97\textwidth]{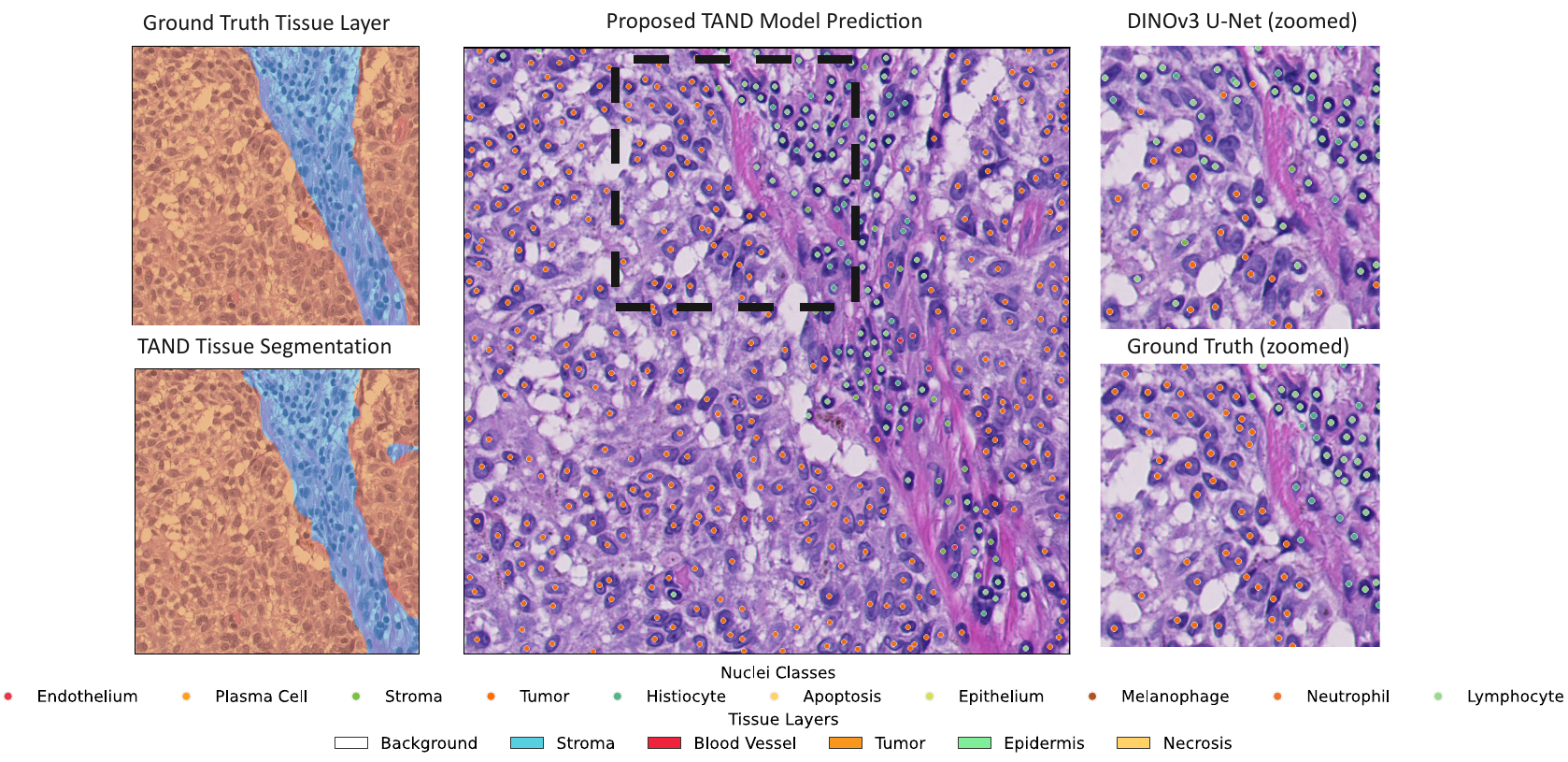}
  \caption{Qualitative visualization of TAND's predictions. \textbf{(Center)} TAND's nuclei overlay on a test sample. \textbf{(Right Column)} Zoomed-in (dashed box) comparing the DINOv3 U-Net baseline (top) and Ground Truth (bottom). \textbf{(Left Column)} Corresponding ground truth (top) and TAND's predicted (bottom) tissue segmentation.}
  \label{fig:visualisation}
\end{figure*}

\subsection{Comparative Evaluation}
TAND requires only point annotations and tissue masks, whereas HoVerNet and HoVerNeXt(baseline method of PUMA dataset~\cite{schuiveling2024puma}) depend on dense nuclear masks, substantially increasing annotation cost. For fair comparison, we extract centre points from their predicted instance masks and apply identical evaluation. Under the same splits in Table~\ref{tab:table1}, TAND achieves the highest Detection F1 and macro-F1 despite weaker supervision. Compared to point-supervised CenterNet~\cite{centernet}, TAND demonstrates superior performance through DINOv3 pretraining and tissue conditioning. Against mask-supervised methods, TAND surpasses HoVerNet across all cell types and outperforms ConvNeXt-based HoVerNeXt (macro-F1: 0.398 vs. 0.374). Crucially, as shown in Fig.~\ref{fig:perclassf1}, improvements are most pronounced in tissue-dependent classes—\textbf{epithelium} (+0.371 F1 over DINOv3), \textbf{endothelium} (+0.298), \textbf{stroma} (+0.199)—validating our tissue-aware design. For tissue segmentation, TAND substantially outperforms U-Net (DICE: 0.545 vs. 0.274).

\subsection{Ablation Study}
To isolate the contribution of tissue conditioning, we remove the FiLM branch (Table~\ref{tab:table1}, ``DINOv3-U-Net''), leaving only the detection–classification backbone. This ablation yields macro-F1 of 0.196 versus 0.398 for full TAND, confirming that multi-scale Spatial-FiLM is critical rather than incremental. Remarkably, tissue conditioning also improves detection F1 (0.734 vs. 0.614) despite operating solely on the classification stream, indicating that tissue-aware features enhance localization by sharpening centre responses and suppressing background peaks. Multi-scale injection at 1/16, 1/8, and 1/4 resolutions maintains fine cellular detail while enforcing cross-scale consistency, yielding complementary gains in both nuclei typing and localization under point supervision.

\subsection{Visualisation Result}
Qualitative results (Fig.~\ref{fig:visualisation}) confirm TAND's benefits. In the zoomed-in region (right column), the DINOv3-Unet baseline misclassifies numerous stromal nuclei as tumour. TAND, leveraging the tissue probability map (left column), correctly classifies these nuclei, resulting in a cleaner and more coherent classification map.

\section{Discussion and Limitations}
Tissue conditioning resolves ambiguities inherent to point supervision while maintaining fast localization. DINOv3 pretraining supplies transferable dense features, explaining TAND's advantage over HoVerNeXt despite similar architectures. A limitation arises when minority tissue layers occupy small regions: downsampling to $16 \times 16$ attenuates tissue layer with small area, reducing contextual benefits for nuclei within those layers. Our residual FiLM design partially mitigates this by adapting features only where tissue evidence is confident. Additional limitations include sensitivity to rare phenotypes, potential propagation of tissue segmentation errors, and domain shift across laboratories. Future work will explore uncertainty-aware gating, class-specific modulation, and multi-magnification conditioning.

\section{Conclusions}
TAND couples a DINOv3 encoder--decoder with a frozen Virchow-2 encoder tissue segmentation branch and applies multi-scale FiLM to the classification stream only. On PUMA it delivers the best detection F1 and macro-F1 among strong baselines while requiring only point annotations plus tissue masks, thus improving the accuracy--to--annotation-cost trade-off. The results show that explicit tissue context, injected with a lightweight and decoupled mechanism, is an effective route to reliable point-supervised nuclei detection and classification suitable for clinical workflows.

\section{Compliance with ethical standards}
\label{sec:ethics}
This study used the publicly available PUMA dataset~\cite{schuiveling2024puma}. Ethical approval was not required as confirmed by the open access licence.

\section{Acknowledgments}
\label{sec:acknowledgments}
Kesi Xu gratefully acknowledges the full PhD scholarship provided by GlaxoSmithKline (GSK) and the Department of Computer Science, University of Warwick, for supporting this research. The authors declare no conflicts of interest.

\bibliographystyle{IEEEbib}
\bibliography{refs}
\end{document}